\documentclass[10pt,twocolumn,letterpaper]{article}

\usepackage[pagenumbers]{cvpr} 

\usepackage{times}
\usepackage{epsfig}
\usepackage{graphicx}
\usepackage{amsmath}
\usepackage{amssymb}
\usepackage{multicol}
\usepackage{multirow}
\usepackage{latexsym}
\usepackage{pifont}
\usepackage{fdsymbol}
\usepackage{marvosym}
\usepackage{flushend}
\usepackage{mathtools}
\usepackage[accsupp]{axessibility} 

\AtBeginEnvironment{tablenotes}{\footnotesize}

\definecolor{cvprblue}{rgb}{0.21,0.49,0.74}
\usepackage[pagebackref,breaklinks,colorlinks,allcolors=cvprblue]{hyperref}

\newcommand{\boldback}[1]{\textbf{\textcolor{black}{\colorbox{cyan!20}{#1}}}}

\definecolor{haoblue}{RGB}{0,112,172}
\definecolor{haoorange}{RGB}{237,125,49}
\usepackage{multirow}

\def\paperID{6914} 
\def\confName{CVPR}
\def\confYear{2025}

\title{DIR: Retrieval-Augmented Image Captioning with Comprehensive Understanding}

\author{Hao Wu\textsuperscript{1,2 $\clubsuit$},\quad Zhihang Zhong\textsuperscript{2 \Letter},\quad Xiao Sun\textsuperscript{2 \Letter} \\
\textsuperscript{1} University of Science and Technology of China~\quad
\textsuperscript{2} Shanghai Artificial Intelligence Laboratory\\
{\tt\small wuhao@mail.ustc.edu.cn,~\{zhongzhihang, sunxiao\}@pjlab.org.cn}
}

\begin{document}
\maketitle
\let\thefootnote\relax\footnotetext{$\clubsuit$ Work done as an intern at Shanghai AI Laboratory.}
\let\thefootnote\relax\footnotetext{\Letter\ Co-corresponding authors.}
\begin{abstract}

Image captioning models often suffer from performance degradation when applied to novel datasets, as they are typically trained on domain-specific data. To enhance generalization in out-of-domain scenarios, retrieval-augmented approaches have garnered increasing attention. However, current methods face two key challenges: (1) image features used for retrieval are often optimized based on ground-truth (GT) captions, which represent the image from a specific perspective and are influenced by annotator biases, and (2) they underutilize the full potential of retrieved text, typically relying on raw captions or parsed objects, which fail to capture the full semantic richness of the data. In this paper, we propose \textbf{D}ive \textbf{I}nto \textbf{R}etrieval (\textbf{DIR}), a method designed to enhance both the image-to-text retrieval process and the utilization of retrieved text to achieve a more comprehensive understanding of the visual content. Our approach introduces two key innovations: (1) diffusion-guided retrieval enhancement, where a pretrained diffusion model guides image feature learning by reconstructing noisy images, allowing the model to capture more comprehensive and fine-grained visual information beyond standard annotated captions; and (2) a high-quality retrieval database, which provides comprehensive semantic information to enhance caption generation, especially in out-of-domain scenarios. Extensive experiments demonstrate that DIR not only maintains competitive in-domain performance but also significantly improves out-of-domain generalization, all without increasing inference costs.

\end{abstract}    
\section{Introduction}
\label{sec:intro}
\begin{figure}[tb]
    \centering
    \includegraphics[width=\linewidth]{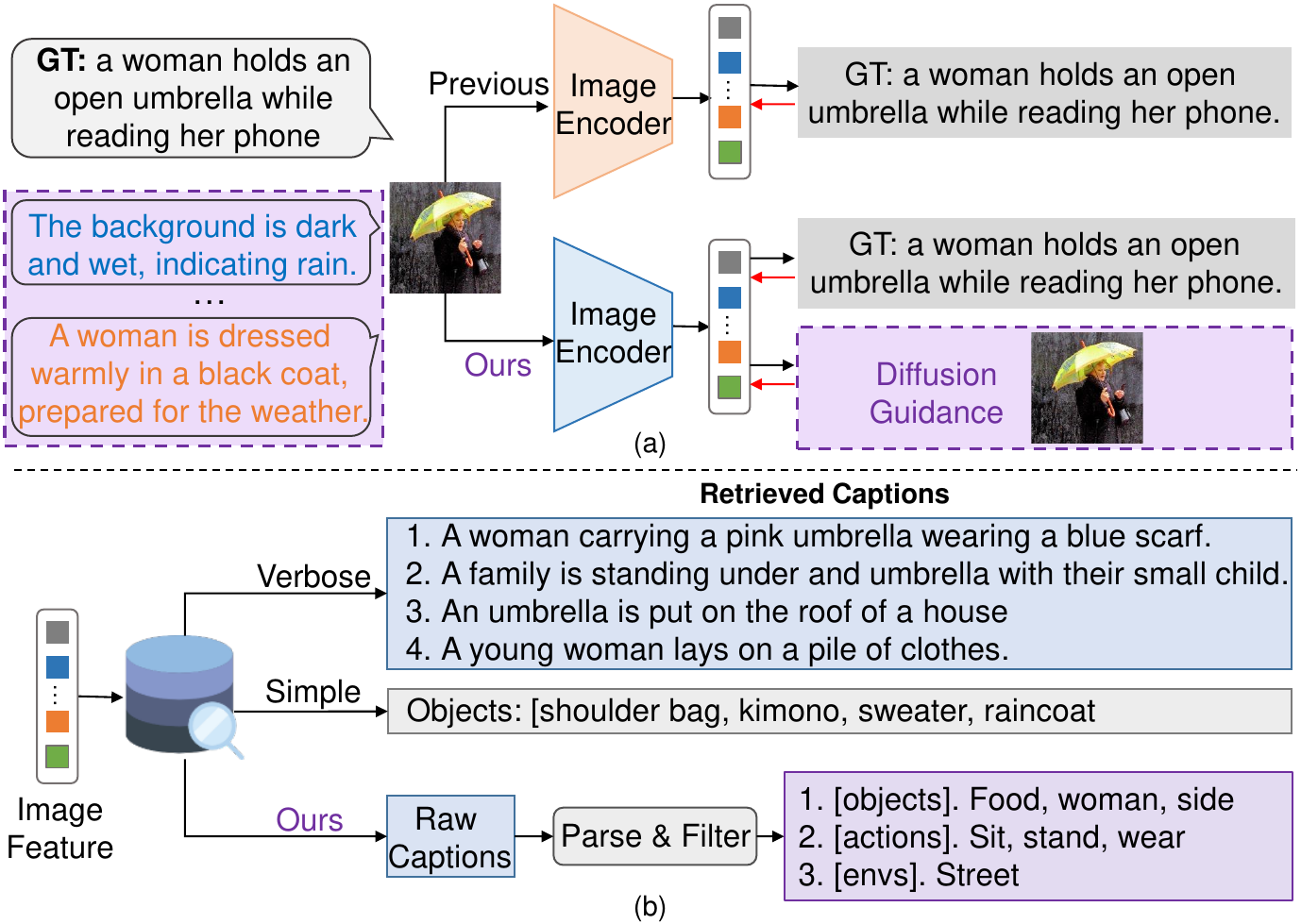}
    \caption{Comparison of our method with previous retrieval-augmented image captioning approaches. (a) \textbf{Diffusion-Guided Retrieval Enhancement:} An image can be described from multiple valid perspectives. Previous methods optimize retrieval features to predict only GT captions, ignoring other perspectives  (highlighted in the left purple dotted box). In contrast, our approach leverages additional diffusion guidance to ensure that image features capture both GT captions and the inherent content of the image, enabling the inclusion of alternative descriptive perspectives.  (b) \textbf{High-Quality Retrieval Database:} Previous methods often rely on raw captions or parsed objects, which are either verbose or overly simplistic. Our approach uses a diverse retrieval text database that captures a broader range of image aspects, including objects, actions, and environments, leading to more contextually rich and accurate captions. \textbf{Please see \Cref{fig:rt_feat} and \Cref{fig:rt_cmp} for examples illustrating the results with and without these methods.}}
    \label{fig:idea}
\end{figure}

Large language model (LLM)-based image captioning has emerged as a powerful approach by integrating pretrained image encoders with LLMs via a projection module, effectively bridging visual and textual representations. Models such as BLIP-2~\cite{Li2023BLIP2BL} and MiniGPT4~\cite{Zhu2023MiniGPT4EV} exemplify this architecture, where the image encoder extracts visual features, and the projection module adapts these features for processing by the LLM. This combination enables the generation of coherent and contextually relevant captions. However, even though this approach takes advantage of pretrained components, the model’s performance is often tailored to the specific in-domain data used during fine-tuning, resulting in strong performance on familiar datasets. In contrast, when applied to novel or out-of-domain data, these models often perform poorly. This limitation arises because the fine-tuning process may not adequately capture the diversity of visual content across different domains.


To improve image captioning, researchers have adopted retrieval-augmented generation (RAG)~\cite{guu2020rag1, lewis2020rag2, asai2024selfrag}, which enhances caption quality by incorporating external, relevant text. RAG allows models to retrieve additional information during inference, enriching captions with context, especially in out-of-domain scenarios. While RAG has proven effective in tasks like question answering and document generation, its use in image captioning remains limited. Existing methods often rely on simple captions or parsed objects, neglecting the rich semantic context that high-quality retrieved text can offer. This highlights the need for an improved retrieval process that better aligns visual features with relevant external information, ultimately enhancing captioning performance.

Optimizing the retrieval process is crucial for improving image captioning, particularly when models are evaluated on out-of-domain or novel datasets, where traditional methods often underperform. A key limitation of existing retrieval-based captioning models is that the visual features for retrieval are typically optimized to predict GT captions, which are biased by annotators' preferences and represent the image from a specific perspective. As shown in \Cref{fig:idea}(a), the image can be described by multiple valid captions (highlighted in the left purple dotted box) beyond the GT caption. However, the retrieval features are trained to focus only on the GT description, leading to a limited representation of the image. This narrow focus results in retrieving only captions closely related to the GT, while ignoring other relevant descriptions that could provide valuable context from different perspectives. Such a bias limits the diversity of retrieved captions, misaligning the retrieval features with the full richness of the image content and reducing performance, especially on diverse or out-of-domain datasets.


Another critical issue in image captioning is the underutilization of retrieved text during caption generation.  Existing methods typically rely on raw captions~\cite{Ramos2022SmallcapLI} or parsed object categories~\cite{Li2023EvcapRI} for contextual information, but these approaches fail to provide a comprehensive range of retrieved text. As shown in \Cref{fig:idea}(b), raw captions often result in overly detailed and specific descriptions, making them unsuitable for different images. On the other hand, relying solely on object categories oversimplifies the context by omitting key visual elements such as actions and environments. These limitations prevent the model from retrieving comprehensive text, which is crucial for generating rich and contextually accurate captions.

To address these challenges, we propose a novel approach to enhance retrieval-augmented image captioning for a more comprehensive understanding of images. Our method introduces two key components: (1) diffusion-guided retrieval enhancement, which leverages a pretrained text-to-image generative diffusion model to improve the retrieval process. Rooted in the philosophy of analysis-by-synthesis, this approach recognizes that image generation inherently requires a compact and holistic analysis of an image’s features. By using image features as conditions for the frozen diffusion model,  we ensure the image encoder learns a more condensed and comprehensive representation. As shown in \Cref{fig:idea}(a), unlike previous methods that rely solely on GT captions for feature learning, our approach incorporates diffusion guidance alongside GT annotations in a self-supervised manner. This reduces the dependency on GT captions and encourages the encoder to capture richer, more comprehensive representations that transcend the limitations of GT labels, better aligning with a broader range of visual contexts.
These comprehensive representations offer a more complete depiction of the image, which is particularly advantageous for retrieval tasks in out-of-domain scenarios. (2) 
A high-quality retrieval database is designed to provide comprehensive image descriptions. As illustrated in \Cref{fig:idea}(b), this database encompasses various aspects of an image, such as objects, actions, and environments, enabling the model to retrieve text that fully captures the contextual richness of the visual content. By categorizing raw sentences into distinct attributes, the retrieved text can be more effectively integrated and adapted to novel datasets. This ensures that the retrieved information is fully utilized during caption generation, resulting in more accurate and contextually comprehensive captions. 

Through these innovations, our method overcomes the limitations of existing retrieval-augmented captioning models. Extensive experiments demonstrate that our framework not only maintains strong in-domain performance but also significantly improves out-of-domain generalization, showcasing the effectiveness of diffusion-guided retrieval enhancement and the high-quality retrieval database in generating comprehensive, contextually accurate captions.

\section{Related Work}
\label{sec:relate}

\subsection{Image Captioning}
Image captioning aims to generate descriptive text based on the content of an image. Traditional methods~\cite{Karpathy2014DeepVA, Anderson2017BottomUpAT} primarily employ encoder-decoder frameworks, where convolutional neural networks (CNNs) encode visual features and recurrent neural networks (RNNs)~\cite{Sutskever2011rnncap} or transformer~\cite{vaswani2017attention} decode these features into natural language captions. As the field has progressed, pretrained models have become increasingly important, with models like BLIP-2~\cite{Li2023BLIP2BL}, MiniGPT4~\cite{Zhu2023MiniGPT4EV} and Flamingo~\cite{alayrac2022flamingo} showcasing the effectiveness of integrating LLMs with pretrained image encoders. These models use a projector module or perceiver resampler to align visual and textual representations, allowing the LLM to generate fluent captions from image features. However, they often rely on large-scale image-text pairs for training and struggle with generalizing to out-of-domain data.

\subsection{Retrieval-Augmented Generation}

RAG~\cite{guu2020rag1, lewis2020rag2, asai2024selfrag} has gained prominence in natural language processing (NLP) for improving the performance of LLMs by incorporating external knowledge during inference. Unlike traditional LLMs, which rely solely on pretrained parameters, RAG models dynamically retrieve relevant information from large external corpora, enhancing both the accuracy and contextual relevance of their outputs.

RAG has also been applied to multimodal tasks such as image and video captioning. In image captioning, retrieval-augmented approaches improve the descriptive quality of captions by leveraging relevant textual data based on visual content. For example, SmallCap~\cite{Ramos2022SmallcapLI} builds a database of image-text pairs to generate captions conditioned on both image features and retrieved text. EXTRA~\cite{Ramos2023extra} and Re-ViLM~\cite{Yang2023Revilm} retrieve captions by comparing the similarity between query image features and those stored in retrieval memory. ViECap~\cite{Fei2023viecap} uses a CLIP-based entity classifier to retrieve relevant entities, boosting zero-shot captioning in out-of-domain scenarios. RA-CM3~\cite{Yasunaga2023Racm3} extends retrieval to a multimodal context by retrieving both text and images using a pretrained CLIP model. MeaCap~\cite{Zeng2024MeaCapMZ} uses a retrieval-then-filter strategy, selecting relevant concepts and refining the caption iteratively with a language model for zero-shot captioning. EVCap~\cite{Li2023EvcapRI} reduces redundancy in retrieved captions by using an external memory that stores image embeddings and object names. In video captioning, R2A~\cite{Pan2023RetrievingtoAnswerZV} enhances retrieval-augmented generation by incorporating frame-level similarity and temporal order for more coherent descriptions. CM$^2$~\cite{Kim2024cm2} applies retrieval-augmented generation to dense video captioning by constructing an external memory bank with sentence-level features for segment-level retrieval.

Previous methods have largely concentrated on constructing and improving the retrieval database by refining the retrieved text from various sources. However, they often overlook the enhancement of image features for better retrieval. In contrast, our approach not only refines the utilization of retrieved text but also improves the features used for retrieval, ultimately leading to better retrieval performance and a more comprehensive understanding of the image.

\subsection{Diffusion Model Representation}

Recent research has demonstrated the versatility of diffusion models across a variety of tasks. For example, SCD-Net~\cite{Luo2022scdnet} and Prefix-Diffusion~\cite{Liu2023PrefixdiffusionAL} utilize diffusion transformers for non-autoregressive captioning, leveraging retrieved text and mapped image embeddings as semantic priors. Diffusion Hyperfeatures~\cite{luo2024diffusionhyper} aggregates multi-scale feature maps from text-to-image diffusion models to enhance performance in downstream tasks, while DDPMSeg~\cite{baranchuk2022ddpmseg} improves semantic segmentation by selecting features from specific layers and timesteps in the diffusion process. Both Diffusion-TTA~\cite{Prabhudesai2023DiffusionTTATA} and DIVA~\cite{wang2024diffusionhelp} employ diffusion-guided feature learning with a frozen diffusion model conditioned on image features; however, Diffusion-TTA focuses on test-time adaptation, while DIVA pretrains an image encoder using diffusion loss before transferring to other tasks. Our method also incorporates diffusion guidance, specifically applied to optimize image features for improved retrieval. Additionally, we introduce a diffusion guidance loss alongside the traditional caption loss during training to balance the quality of retrieval and the caption generation. This approach enables the model to leverage better retrieval text as context for caption generation, distinguishing our method from previous approaches in both motivation and execution.
\section{Method}
In this work, we aim to enhance retrieval-augmented image captioning by addressing two key challenges. The first challenge is the bias in image features optimized using GT captions. To mitigate this, we introduce diffusion guidance alongside traditional caption loss, encouraging the image encoder to capture a broader range of visual information for a more comprehensive image representation. The second challenge is the underutilization of retrieved text. Current methods often depend on either overly verbose captions or overly simplified object-based descriptions. To overcome this, we propose a high-quality retrieval text database that includes detailed descriptions of objects, actions, and environments, allowing the model to generate more accurate and contextually rich captions.

\subsection{Network Architecture}
\begin{figure*}[tb]
    \centering
    \includegraphics[width=\linewidth]{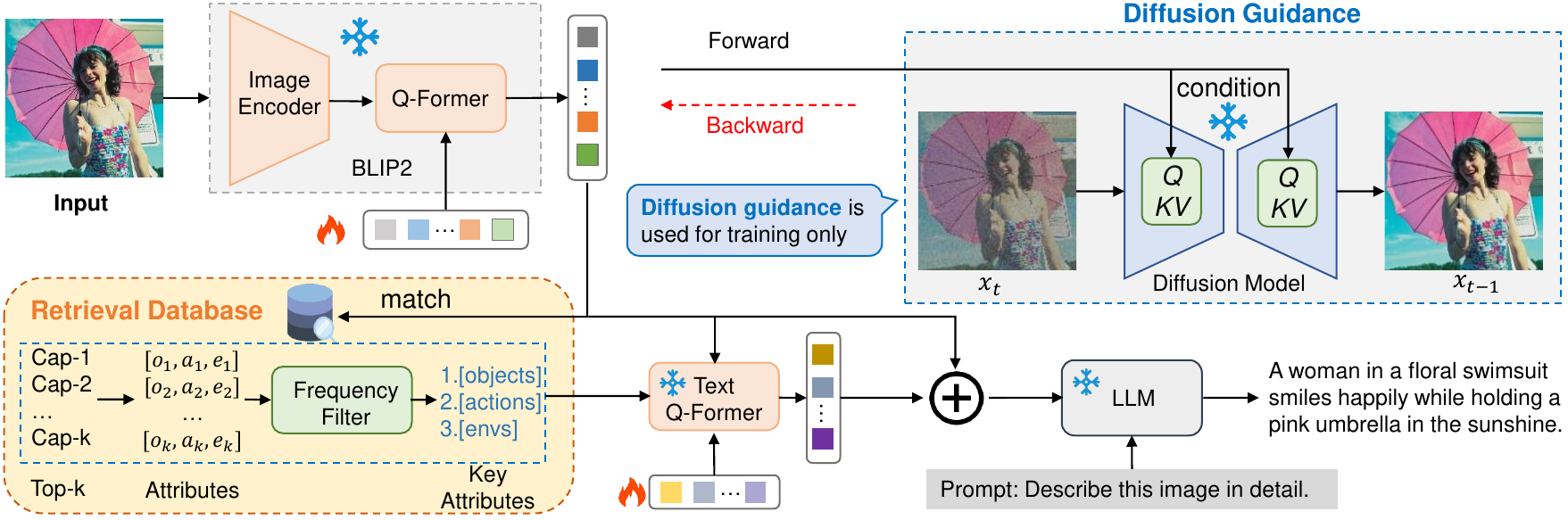}
    \caption{The architecture of our retrieval-augmented image captioning framework employs an image encoder and Q-Former from BLIP2, guided by a pretrained text-to-image diffusion model, to extract comprehensive image features for retrieval. The retrieved text features are fused with the image features through a Text Q-Former, and the combined features are then passed to an LLM for final caption generation.}
    \label{fig:method}
\end{figure*}

Our approach builds on the EVCap framework~\cite{Li2023EvcapRI}, which achieves state-of-the-art performance by integrating LLM-based image captioning with a retrieval database while minimizing training parameters. We extend this foundation by incorporating diffusion-guided retrieval techniques and leveraging a high-quality retrieval database, enhancing the effectiveness of retrieval-augmented image captioning, as illustrated in \Cref{fig:method}. The image encoder is guided by a frozen, pretrained text-to-image diffusion model, enabling it to extract more robust and comprehensive image features. These features are then used to retrieve relevant content from a database containing a broader range of attributes, such as objects, actions, and environments, ensuring richer and more contextually relevant information. A frequency-based filtering mechanism selects the most pertinent terms from the retrieved text. Following the EVCap framework, the filtered text features are fused with the image features using a customized Text Q-Former, designed to extract and integrate the most relevant text features related to the image. The query embeddings for both the Q-Former and Text Q-Former are learnable, optimizing the extraction of comprehensive image features and the most relevant text features, respectively. The image and fused text features are concatenated and passed into a LLM to generate the final caption. 
By combining diffusion-guided retrieval enhancement with a high-quality retrieval database, our method significantly enhances the accuracy and comprehensiveness of caption generation.

\subsection{Diffusion-Guided Retrieval Enhancement}
Our method enhances retrieval-augmented image captioning by using a text-to-image diffusion model to guide the image encoder in learning comprehensive features. These features improve retrieval performance and enable the generation of more accurate captions.

\subsubsection{Diffusion Model Overview}
The Diffusion Model~\cite{Rombach2021latentdiffusion} is a generative framework that creates high-quality images by progressively denoising a noisy input. In the forward process, an image $x_0$ is transformed into a noisy version $x_T$ by adding Gaussian noise over several time steps. In the reverse process, the model learns to recover the original image by predicting and removing the noise introduced at each step. 
The denoising is conditioned on additional information $c$, such as text, semantic maps. The model, denoted as $\epsilon_\theta(x_t, c, t)$, predicts the noise  at each time step, progressively reconstructing the clean image.

The training objective minimizes the difference between the predicted and actual noise $\epsilon$ at each step, as represented by the following loss function:
\begin{equation}
\mathcal{L}_{\text{diffusion}} = \mathbb{E}_{x_0, t, \epsilon \sim \mathcal{N}(0, 1)} \left[ \|\epsilon - \epsilon_\theta(x_t, c, t)\|^2 \right].
\end{equation}
This objective drives the model to effectively learn the reverse diffusion process, progressively refining the noisy inputs back into coherent, high-quality images.



\subsubsection{Feature Learning via Diffusion Guidance}
In our approach, the diffusion model is conditioned on image features instead of textual data or other modalities.  The image encoder $\mathcal{E}$ extracts a feature vector $z = \mathcal{E}(x_0)$ from the input image $x_0$, which  guides the denoising process. By leveraging these image features, the model learns to reconstruct the original image from the noisy version $x_t$, focusing on the holistic visual content rather than biased perspectives of GT caption. As shown in the \textcolor{haoblue}{\textbf{Diffusion Guidance}} block of \Cref{fig:method}, the model uses the noisy image $x_t$, the current timestep $t$, and the extracted image feature $z$ as inputs, formulated as $\epsilon_\theta(x_t, t, z)$. The model is trained in a self-supervised manner with the following loss function:
\begin{equation}
   \mathcal{L}_{\text{denoise}} = \mathbb{E}_{x_0, z, t, \epsilon \sim \mathcal{N}(0, 1)} \left[ \|\epsilon - \epsilon_\theta(x_t, t, z)\|^2 \right].
   \label{eq:denoise_loss}
\end{equation}
By conditioning the diffusion model on image features, our method enables the encoder to capture a holistic image representation without relying solely on ground truth captions. This comprehensive understanding improves text retrieval from diverse perspectives, enhancing retrieval performance and enabling more accurate, contextually rich captions. Importantly, the diffusion guidance is employed exclusively during training, without introducing any additional inference costs.

\subsection{High-Quality Retrieval Database}
In our method, we construct a comprehensive retrieval database that significantly improves the retrieval process by incorporating a broader range of attributes. Unlike previous approaches that focus solely on raw captions or object categories, our database decomposes raw captions and integrates additional attributes like objects, actions, and environments. This enriched representation provides a more complete view of the visual content, enhancing both the quality of the retrieved text and the contextual information for caption generation. As a result, the model generates more accurate and contextually relevant captions, especially in out-of-domain scenarios.

A key aspect of our approach is the parsing and categorization of retrieved text. In MeaCap~\cite{Zeng2024MeaCapMZ}, a zero-shot image captioning method, raw captions are decomposed into subject-predicate-object triplets, which are then merged, filtered, and refined to generate the final sentence. This fixed triplet structure is effective in zero-shot settings, where caption generation relies solely on retrieved texts, and the triplet format provides guidance. However, we argue that this approach is less suitable for training-based methods. In training scenarios, the model can leverage detailed and comprehensive image features, making the rigid triplet structure unnecessary. Additionally, imposing such priors can introduce noise, as triplets may not always appear in captions, leading to suboptimal results. Thus, as shown in the \textcolor{haoorange}{\textbf{Retrieval Database}} block of \Cref{fig:method}, we adopt a more flexible decomposition strategy, breaking sentences into multiple attributes. While previous works~\cite{Zeng2024MeaCapMZ,Li2023EvcapRI,wang2022vidil} focus primarily on objects and actions, we suggest that the environment, such as a kitchen or playground, provides valuable context and serves as a powerful indicator to help infer the overall scene and better understand the image. To improve the quality and contextual relevance of the generated captions, we introduce a frequency-based filtering mechanism to remove noisy or irrelevant terms. Rather than using a fixed threshold, we apply a top-$n$ selection based on term frequency within each category (objects, actions, and environments), ensuring that only the most relevant attributes are retained for further processing. Formally, the filtering process is expressed as:
\begin{equation}
\mathbf{C}_{\text{filtered}} = \{ c_i \mid \text{freq}(c_i) \in \text{top-}n \},
\label{eq:freq}
\end{equation}
where $\mathbf{C}_{\text{filtered}}$ represents the set of filtered attributes, and $c_i$ denotes individual components (objects, actions, or environments). After filtering, the selected attributes are organized into a structured sentence through a soft prompt like \textit{``[images] and [actions] in [scenes]"}. This sentence, now containing the filtered attributes, is then sent to a BERT~\cite{Devlin2019bert} encoder to generate a contextualized text representation.

For the image features in the retrieval database, we utilize the raw features extracted by EVA-CLIP~\cite{Fang2022eva}. These raw features are prioritized over those generated by an image encoder trained with diffusion guidance due to the critical requirement for retrieval features to be closely aligned with their corresponding text. Effective image-text matching relies on this alignment to ensure that the retrieved information is relevant and informative. The EVA-CLIP model, which is trained using contrastive loss, has been specifically designed to optimize the similarity between images and their associated textual descriptions. This results in a robust representation that ensures images are accurately mapped to corresponding text, making raw CLIP features particularly well-suited for constructing the retrieval database. 
\subsection{Model Training}


Our model employs a joint optimization strategy to enhance both retrieval performance and caption generation accuracy. It combines standard next-token prediction with diffusion-guided retrieval enhancement, balancing these components to improve retrieval and utilize the retrieved text context for better captioning. The training process uses two main loss functions: diffusion denoising loss and caption loss. Diffusion-guided retrieval enhancement is driven by the denoising loss, where the image encoder learns to reconstruct original images from noisy inputs based on image features. This process allows the model to capture comprehensive image representations that improve retrieval performance, as defined in \Cref{eq:denoise_loss}. The caption loss ensures high-quality captions by leveraging both image and retrieved text features, calculated using cross-entropy between the predicted caption and the GT captions, as shown in the following equation:
\begin{equation}
\mathcal{L}_{\text{caption}} = - \sum_{i=1}^{N} \log P_{\theta}(w_i | w_{1:i-1}, \mathbf{X}),
\label{eq:lmloss}
\end{equation}
where \( w_i \) denotes the $i$-th  word of GT caption, and \( \mathbf{X} \) represents the concatenated representation of image features and the contextualized text features obtained from the retrieval process.  The total training loss combines both losses to improve retrieval performance and ensure accurate caption generation. The total loss is expressed as: 
\begin{equation}
\mathcal{L}_{\text{total}} = \mathcal{L}_{\text{caption}}  + \lambda \mathcal{L}_{\text{denoise}},
\end{equation}
where \( \lambda \) controls the relative weighting of the two losses. This joint training strategy optimizes both the retrieval process and caption generation, resulting in improved retrieval quality and more accurate captioning.

\section{Experiments}

In this section, we outline the experimental setup and report the model's performance on both in-domain and out-of-domain datasets, emphasizing its ability to generalize to out-of-domain data while maintaining strong performance on in-domain data. Additionally, we present qualitative analysis and  ablation analysis to further evaluate the effectiveness of our approach.

\subsection{Experimental Settings}
\noindent\textbf{Datasets.} 
We conduct experiments on three datasets: COCO~\cite{Lin2014coco}, Flickr30k~\cite{Plummer2015Flickr30kEC}, and NoCaps~\cite{Agrawal2019nocaps}. COCO is used for training, with evaluation carried out on the COCO test set using the Karpathy split~\cite{Karpathy2014DeepVA}, as well as on Flickr30k and NoCaps. NoCaps includes in-domain, near-domain, and out-of-domain subsets, enabling us to assess model performance under varying domain shifts and its generalization to both familiar and unseen domains.

\vspace{0.5mm}
\noindent\textbf{Implementation.}
Our model builds on the EVCap framework, utilizing a Vision Transformer (ViT)~\cite{dosovitskiy2020visiontransformer} and Q-Former from BLIP2~\cite{Li2023BLIP2BL} as the image encoder along with a text Q-Former to integrate text and image features.  A BERT~\cite{Devlin2019bert} model processes the retrieved textual attributes, while the latent diffusion model, Stable Diffusion v1.4~\cite{Rombach2021latentdiffusion}, guides the image encoder's learning during training. Final caption generation is performed by Vicuna-13B~\cite{chiang2023vicuna}, which receives the concatenated image and fused text features.

\vspace{0.5mm}
\noindent\textbf{Evaluation.} 
We evaluate our model using standard image captioning metrics, including BLEU@4 (B4)~\cite{Papineni2002BleuAM}, METEOR (M)~\cite{banerjee2005meteor}, CIDEr (C)~\cite{Vedantam2014CIDErCI}, and SPICE (S)~\cite{anderson2016spice}, across multiple datasets. The evaluation includes two primary comparisons: in-domain performance on the COCO test set and the in-domain, near-domain subset of NoCaps, and out-of-domain performance on Flickr30k, along with the out-of-domain and overall subsets of NoCaps. This structure demonstrates the model’s strong in-domain results and effective generalization to out-of-domain scenarios.

\subsection{In-Domain and Out-of-Domain Performance}


\begin{table}[tb]
\resizebox{\columnwidth}{!}{%
\begin{tabular}{l|cc|cc|cccc}
\toprule
\multirow{3}{*}{\textbf{Method}} & \multicolumn{2}{c|}{\textbf{Train}} & \multicolumn{2}{c|}{\textbf{Flickr30k}} & \multicolumn{4}{c}{\textbf{NoCaps Val}} \\ \cmidrule(l){2-9} 
                                   & \multicolumn{1}{c|}{Data} & Para. & \multicolumn{2}{c|}{Test}  & \multicolumn{2}{c|}{Out-domain} & \multicolumn{2}{c}{Overall} \\
                                   &      &      & \multicolumn{1}{|c}{C} & S & \multicolumn{1}{|c}{C} & S & \multicolumn{1}{|c}{C} & S \\ \midrule
\multicolumn{9}{l}{\textbf{Heavyweight training methods}} \\
BLIP~\cite{Li2022BLIPBL}  &  129M    & 446B   & - & -  & 115.3 & 14.4  &  113.2 & 14.8  \\ 
BLIP2~\cite{Li2023BLIP2BL}  &  129M    & 1.2B & - & -  & 124.8 & 15.1  &  121.6 & 15.8 \\
ViECap~\cite{Fei2023viecap}  &  COCO    & 124M & 47.9 & 13.6  & 65.0 & 8.6  &  66.2 & 9.5  \\ \midrule
\multicolumn{9}{l}{\textbf{Lightweight training methods}} \\
MiniGPT4~\cite{Zhu2023MiniGPT4EV}  &  5M    & 3.94M & 78.4 & 16.9  & 110.8 & 14.9  &  108.8 & 15.1  \\
ClipCap~\cite{Mokady2021ClipCapCP}&  COCO   & 43M   & - & -  &     49.1    & 9.6  &  65.8  & 10.9  \\
SmallCap~\cite{Ramos2022SmallcapLI}& COCO   &  7M   & 60.6  & -    & - &  - & - & -  \\
EVCap~\cite{Li2023EvcapRI}         &  COCO & 3.97M & 84.4   & 18.0    &   116.5 & 14.7  & 119.3 & 15.3  \\
\textbf{DIR (ours)}                               &  COCO & 3.97M &  \boldback{85.7}  & \boldback{18.2}   &  \textbf{118.7} & \boldback{15.1}  & \textbf{120.7} & \textbf{15.5}  \\ \bottomrule
\end{tabular}%
}
\caption{Out-of-domain performance comparison on Flickr30k and NoCaps (out-domain and overall). The best performance among lightweight methods is highlighted in bold, with a light blue background  added when it also surpasses all heavyweight models.}
\label{tab:outdomain}
\end{table}
\noindent\textbf{Out-of-Domain Performance.} 
\Cref{tab:outdomain} presents a comparison of our model with both lightweight and heavyweight methods on out-of-domain datasets, including Flickr30k, as well as the out-of-domain and overall subsets of NoCaps. Despite being trained solely on the COCO dataset, our model consistently outperforms all lightweight approaches across every metric, thanks to the diffusion-guided retrieval enhancement and a high-quality retrieval database. Despite having the same training data and parameter count as EVCap, our model leverages these innovations to achieve superior performance. Remarkably, our approach competes with or even surpasses heavyweight models, which are typically trained on larger datasets and have more learnable parameters. This highlights the strong generalization capability of our method, enabling it to handle complex out-of-domain scenarios effectively with minimal additional training resources.


\begin{table}[tb]
\centering
\resizebox{0.95\columnwidth}{!}{%
\begin{tabular}{c|cccc|cccc}
\toprule
\multicolumn{1}{c|}{\multirow{3}{*}{\textbf{Method}}} & \multicolumn{4}{c|}{\textbf{COCO}} & \multicolumn{4}{c}{\textbf{NoCaps Val}} \\
\multicolumn{1}{c}{} & \multicolumn{4}{|c}{Test} & \multicolumn{2}{|c}{In-domain} & \multicolumn{2}{|c}{Near-domain} \\
\multicolumn{1}{c|}{}                        & B4    & M    & C    & S   & \multicolumn{1}{c}{C}  & S & \multicolumn{1}{|c}{C}  & S                                     \\ \midrule
ClipCap~\cite{Mokady2021ClipCapCP} & 33.5  & 27.5  & 113.1 & 21.1              & 84.9 & 12.1 & 66.8 & 10.9 \\
SmallCap~\cite{Ramos2022SmallcapLI} & 37.0  & 27.9  & 119.7 & 21.3              & - & - & - & - \\
EVCap~\cite{Li2023EvcapRI}          & 41.5  & 31.2  & 140.1 & 24.7  & \textbf{111.7}  & 15.3 & 119.5 & 15.6 \\
\textbf{DIR (ours)}                                   &  \textbf{41.7} &  \textbf{31.5} & \textbf{140.9} & \textbf{25.0} & \textbf{111.7} & \textbf{15.5} & \textbf{120.6} & \textbf{15.8}                                   \\ \bottomrule
\end{tabular}%
}
\caption{In-domain performance comparison on the COCO test set and NoCaps validation set (in-domain and near-domain). The best performance is highlighted in bold.}
\label{tab:indomain_comparison}
\end{table}

\vspace{0.5mm}
\noindent\textbf{In-Domain Performance.}
\Cref{tab:indomain_comparison} compares our method with several prior lightweight models, particularly EVCap, which improves captioning performance by utilizing a larger language decoder, Vicuna-13B, compared to models like SmallCap and CLIPCap that use GPT-2. Building on EVCap's strong performance, our model employs the same image encoder and LLM as EVCap. Both our model and EVCap are trained on the same dataset with identical learnable parameter counts. However, our method introduces two key innovations: diffusion-guided retrieval enhancement and a high-quality retrieval database. Evaluation on COCO and the in-domain and near-domain subsets of NoCaps shows that our approach consistently outperforms these models across most metrics. While designed to improve out-of-domain performance, our method also maintains and even enhances in-domain results without adding extra trainable parameters.

\subsection{Qualitative Analysis}
To evaluate the effects of the proposed modules, we present a qualitative analysis of the model's performance, with a particular focus on the effects of diffusion-guided retrieval enhancement and the high-quality retrieval database on out-of-domain data.

\begin{figure}[tb]
    \centering
    \includegraphics[width=\linewidth]{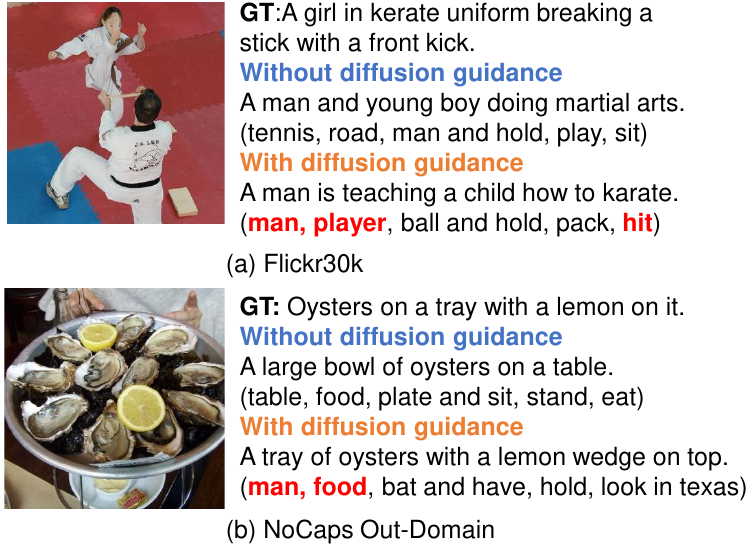}
    \caption{Qualitative comparison of model performance with and without diffusion guidance. The relevant retrieval results are highlighted in bold and red.}
    \label{fig:rt_feat}
\end{figure}

\vspace{0.5mm}
\noindent\textbf{Retrieval Features with and without Diffusion Guidance.}
We present a qualitative analysis of model performance with and without diffusion guidance. In \Cref{fig:rt_feat}(a), the model utilizing diffusion-guided image features retrieves relevant context, such as ``man", ``player" and ``hit", aiding the LLM in generating an accurate caption. However, with raw feature retrieval, the key action ``hit" is not retrieved, limiting the model's ability to predict the correct activity, such as ``karate." In \Cref{fig:rt_feat}(b), both raw image features and diffusion-guided features retrieve relevant context, but the model with diffusion guidance outperforms the other by correctly identifying ``lemon at the top of oysters." This suggests that diffusion guidance captures finer details, even when such information is not directly retrieved. We hypothesize that the denoising process in diffusion guidance encourages the image encoder to focus on the entire image, resulting in a more comprehensive understanding of the visual content. In contrast, when trained with caption supervision alone, the encoder tends to focus on more prominent objects, potentially overlooking finer details.

\begin{figure}[tb]
    \centering
    \includegraphics[width=\linewidth]{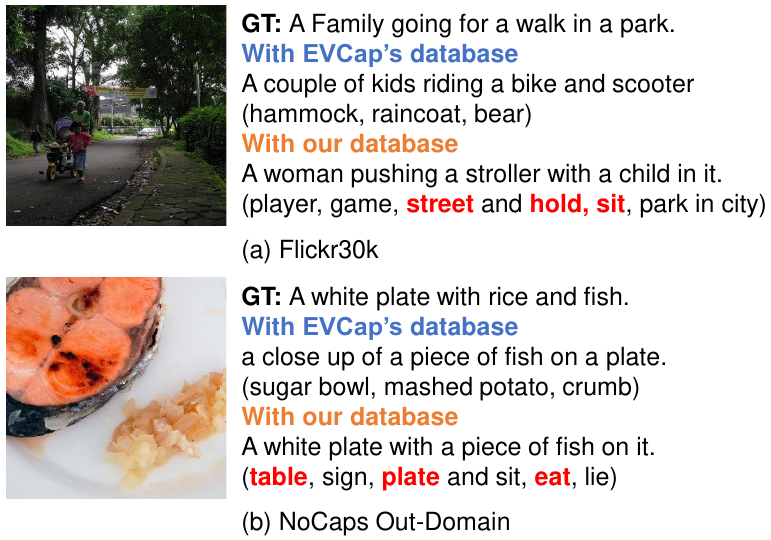}
    \caption{Qualitative comparison of model predictions using EVCap's retrieval database and our proposed retrieval database. The relevant retrieval results are highlighted in bold and red.}
    \label{fig:rt_cmp}
\end{figure}

\vspace{0.5mm}
\noindent\textbf{Retrieval Results with Different Retrieval Databases.}
We present a qualitative analysis of model performance using both EVCap's retrieval database and our proposed retrieval database. \Cref{fig:rt_cmp} visualizes the model's predictions on two examples from the Flickr30k test set and the NoCaps out-of-domain validation set. Our model, leveraging the proposed database, captures not only object-related context but also action terms like ``hold", ``sit", and ``eat" as well as environmental terms such as ``city", enabling the LLM to generate more accurate and contextually rich captions. This emphasizes the importance of integrating both object and action context for high-quality captioning. In contrast, EVCap's retrieval database relies on terms from the limited LVIS dataset, which offers a narrower range of object descriptions. This results in less diverse retrieval terms, limiting the guidance available for caption generation. For the same input image, significant differences are observed in the retrieved results from the two databases. EVCap’s retrieval often focuses on isolated object descriptions, whereas our database captures a more holistic view, including objects, dynamic actions and environmental context. This broader set of attributes enables our model to generate more comprehensive and contextually relevant captions by retrieving information that reflects the full scene, rather than just individual objects. This comparison highlights the crucial role of a comprehensive retrieval database in improving caption quality and enhancing contextual understanding.

\subsection{Ablation Analysis}

\noindent\textbf{Effect of the Retrieval Database.} 
To thoroughly evaluate the effect of the retrieval database on model performance, we compare several configurations: a model without retrieval, one using the EVCap-defined retrieval database, and model with our retrieval database. As seen in \Cref{tab:effetc_retrieval}, the use of EVCap’s retrieval database improves most metrics, although it results in a slight drop in the Flickr30k CIDEr score. In contrast, when using our proposed retrieval database, the model demonstrates consistent improvement across all metrics, with more significant gains compared to EVCap’s database. This improvement is largely attributed to the enhanced contextual richness introduced by our retrieval process, which enables the model to generalize better to out-of-domain data.

\begin{table}[tb]
\resizebox{\columnwidth}{!}{%
\begin{tabular}{l|cc|cccccc}
\toprule
\multirow{2}{*}{\textbf{Database}} & \multicolumn{2}{c|}{\textbf{Flickr30k}}  & \multicolumn{2}{c}{\textbf{NoCaps-Near}} & \multicolumn{2}{c}{\textbf{NoCaps-Out}} & \multicolumn{2}{c}{\textbf{NoCaps-All}} \\ \cmidrule(l){2-9} 
                                   & \multicolumn{1}{|c}{C} & S  & \multicolumn{1}{|c}{C} & S & \multicolumn{1}{|c}{C} & S & \multicolumn{1}{|c}{C} & S \\ \midrule
w/o Retrieval         &  84.3   & 18.0 & 118.7 & 15.7  & 117.5 & 14.9  & 119.0 & 15.4  \\
EVCap's        &  84.1   & 18.1  & 119.5 & \textbf{15.8} &   118.1 & 14.9  & 119.9 & \textbf{15.5}  \\
\textbf{Ours  }         &  \textbf{85.7}  & \textbf{18.2} & \textbf{120.6} & \textbf{15.8} &  \textbf{118.7} & \textbf{15.1}  & \textbf{120.7} & \textbf{15.5}  \\ \bottomrule
\end{tabular}%
}
\caption{Comparison of model performance across different retrieval configurations: without retrieval, using the EVCap-defined retrieval database, and using our proposed retrieval database. }
\label{tab:effetc_retrieval}
\end{table}

\vspace{0.5mm}
\noindent\textbf{Effect of Diffusion-Guided Retrieval Enhancement.} 
To assess the effect of diffusion-guided retrieval enhancement on model performance, we compare models performance with and without the diffusion guidance. As shown in \Cref{tab:effetc_diffusion_guidance}, the model that incorporates diffusion-guided retrieval enhancement consistently outperforms the baseline across various metrics, with a particularly noticeable improvement on out-of-domain datasets. This performance boost highlights the ability of diffusion guidance to help the model learn more comprehensive and transferable image features. The significant boost in out-of-domain performance demonstrates the effectiveness of diffusion-guided retrieval enhancement in enhancing the model's generalization and adaptability to diverse, unseen data.

\begin{table}[tb]
\resizebox{\columnwidth}{!}{%
\begin{tabular}{l|cc|cccccc}
\toprule
\multirow{2}{*}{\textbf{Setting}} & \multicolumn{2}{c|}{\textbf{Flickr30k}}  & \multicolumn{2}{c}{\textbf{NoCaps-Near}} & \multicolumn{2}{c}{\textbf{NoCaps-Out}} & \multicolumn{2}{c}{\textbf{NoCaps-All}} \\ \cmidrule(l){2-9} 
                                   & \multicolumn{1}{|c}{C} & S  & \multicolumn{1}{|c}{C} & S & \multicolumn{1}{|c}{C} & S & \multicolumn{1}{|c}{C} & S \\ \midrule
w/o Diffusion guidance         &  83.9   & 18.1 & 119.8 & 15.7  &   117.2 & 14.9  & 119.7 & 15.4  \\
w/ Diffusion guidance         &  \textbf{85.7}  & \textbf{18.2} & \textbf{120.6} & \textbf{15.8} &  \textbf{118.7} & \textbf{15.1}  & \textbf{120.7} & \textbf{15.5}  \\ \bottomrule
\end{tabular}%
}
\caption{Comparison of model performance with and without diffusion guidance on Flickr30k and NoCaps.}
\label{tab:effetc_diffusion_guidance}
\end{table}
\vspace{0.5mm}
\noindent\textbf{Effect of Retrieval Text as an Extra Condition.} We further investigate the impact of using retrieved text as an additional condition in the diffusion guidance, where the condition embedding is derived by averaging the image and retrieved text embeddings. As shown in \Cref{tab:effetc_rt_condition}, our results reveal that incorporating text as a condition leads to reduced performance, especially on out-of-domain datasets. We hypothesize that while the retrieval text may assist in the denoising process, it may also hinder the effective training of the image encoder. Specifically, the encoder is not trained to capture information present in the text features, and the non-differentiable nature of the retrieval process prevents gradients from being backpropagated from the conditioned text embedding to the image encoder. This limitation restricts the image encoder’s ability to learn optimal image representations.

\begin{table}[tb]
\resizebox{\columnwidth}{!}{%
\begin{tabular}{l|cc|cccccc}
\toprule
\multirow{2}{*}{\textbf{Setting}} & \multicolumn{2}{c|}{\textbf{Flickr30k}}  & \multicolumn{2}{c}{\textbf{NoCaps-Near}} & \multicolumn{2}{c}{\textbf{NoCaps-Out}} & \multicolumn{2}{c}{\textbf{NoCaps-All}} \\ \cmidrule(l){2-9} 
                                   & \multicolumn{1}{|c}{C} & S  & \multicolumn{1}{|c}{C} & S & \multicolumn{1}{|c}{C} & S & \multicolumn{1}{|c}{C} & S \\ \midrule
w/ Text condition         &  84.6   & 18.1  & 120.3 & 15.6 &   117.2 & 14.8  & 120.1 & 15.4  \\
w/o Text condition       &  \textbf{85.7}  & \textbf{18.2} & \textbf{120.6} & \textbf{15.8} &  \textbf{118.7} & \textbf{15.1}  & \textbf{120.7} & \textbf{15.5}  \\ \bottomrule
\end{tabular}%
}
\caption{Performance comparison of models with and without retrieved text as an additional condition in the diffusion model on Flickr30k and NoCaps.}
\label{tab:effetc_rt_condition}
\end{table}
\vspace{0.5mm}
\noindent\textbf{Effect of Feature Fusion.} 
To investigate the effects of feature fusion on image features, we compare the model's performance using raw image features, which are the output of the image encoder trained with diffusion guidance, to fused image features. The fused image features are obtained by combining raw image features with retrieved text features using the Text Q-Former. As shown in \Cref{tab:effetc_image_feature}, our experiments indicate that using fused image features results in a performance decrease across both in-domain and out-of-domain datasets. This suggests that while the Text Q-Former reduces noise in the text features through cross-attention, the fusion process may introduce noise to the fused image features from the retrieved text, negatively impacting the fused image representation. Therefore, feature fusion is not always beneficial, and we opt to use raw image features combined with fused text features to achieve better performance.
\begin{table}[tb]
\resizebox{\columnwidth}{!}{%
\begin{tabular}{l|cc|cccccc}
\toprule
\multirow{2}{*}{\textbf{Setting}} & \multicolumn{2}{c|}{\textbf{Flickr30k}}  & \multicolumn{2}{c}{\textbf{NoCaps-Near}} & \multicolumn{2}{c}{\textbf{NoCaps-Out}} & \multicolumn{2}{c}{\textbf{NoCaps-All}} \\ \cmidrule(l){2-9} 
                                   & \multicolumn{1}{|c}{C} & S  & \multicolumn{1}{|c}{C} & S & \multicolumn{1}{|c}{C} & S & \multicolumn{1}{|c}{C} & S \\ \midrule
Fused image feature        &  83.0   & 17.9  & 118.3 & 15.5 & 113.9   & 14.6  & 117.9 & 15.2  \\
Raw image feature        &  \textbf{85.7}  & \textbf{18.2} & \textbf{120.6} & \textbf{15.8} &  \textbf{118.7} & \textbf{15.1}  & \textbf{120.7} & \textbf{15.5}  \\ \bottomrule
\end{tabular}%
}
\caption{Comparison of model performance using raw image features versus fused image features on Flickr30k and NoCaps.}
\label{tab:effetc_image_feature}
\end{table}

\vspace{0.5mm}
\noindent\textbf{Effect of Diffusion Loss Weight $\lambda$.}
We evaluate the effect of the diffusion loss weight  \(\lambda\) on model performance by conducting experiments with various values. As shown in \Cref{tab:effect_weight}, increasing \(\lambda\) generally improves performance, especially for out-of-domain data. A higher \(\lambda\) enables the model to better leverage diffusion-guided retrieval, enriching contextual information and enhancing generalization across diverse scenarios. However, these performance gains are not always consistent. And when \(\lambda\) becomes too large, performance starts to decline, with the tipping point varying across datasets. We hypothesize that an excessive diffusion loss weight may cause the model to over-optimize for denoising, shifting focus away from caption optimization and disrupting the balance between the two objectives, ultimately leading to suboptimal performance.

\begin{table}[tb]
\resizebox{\columnwidth}{!}{%
\begin{tabular}{c|cc|cccccc}
\toprule
\multirow{2}{*}{\textbf{weight}-$\lambda$} & \multicolumn{2}{c|}{\textbf{Flickr30k}}  & \multicolumn{2}{c}{\textbf{NoCaps-Near}} & \multicolumn{2}{c}{\textbf{NoCaps-Out}} & \multicolumn{2}{c}{\textbf{NoCaps-All}} \\ \cmidrule(l){2-9} 
                                   & \multicolumn{1}{|c}{C} & S  & \multicolumn{1}{|c}{C} & S & \multicolumn{1}{|c}{C} & S & \multicolumn{1}{|c}{C} & S \\ \midrule
3         &  84.0   & 18.1 & 119.1 & 15.7   &   117.3 & 14.9  & 119.1 & 15.4  \\
5         &  \textbf{86.2}   & \textbf{18.3}  & 120.3 & \textbf{15.9} &   \textbf{118.9} & \textbf{15.1}  & 120.5 & \textbf{15.6}  \\
7       &  85.7  & 18.2 & \textbf{120.6} & 15.8 &  118.7 & \textbf{15.1}  & \textbf{120.7} & 15.5  \\
9       &  84.9  & \textbf{18.3} & 119.9 & 15.8 &  118.2 & 15.0  & 120.1 & \textbf{15.6}  \\ \bottomrule
\end{tabular}%
}
\caption{ Performance comparison of varying diffusion loss weight $\lambda$ on Flickr30k and NoCaps.}
\label{tab:effect_weight}
\end{table}

\section{Conclusion}

We propose a novel image captioning approach that integrates diffusion-guided retrieval with a high-quality retrieval database. Our model consistently outperforms existing methods in out-of-domain settings while maintaining strong in-domain performance. By leveraging diffusion-guided retrieval and enriched context, it generates more accurate, contextually rich captions and generalizes better to unseen data.
\section*{Acknowledgments}
\begin{flushleft}
This work is partially supported by the National Key R\&D Program of China (NO.2022ZD0160104).
\end{flushleft}
{
    \small
    \bibliographystyle{ieeenat_fullname}
    \bibliography{main}

\begin{thebibliography}{40}
\providecommand{\natexlab}[1]{#1}
\providecommand{\url}[1]{\texttt{#1}}
\expandafter\ifx\csname urlstyle\endcsname\relax
  \providecommand{\doi}[1]{doi: #1}\else
  \providecommand{\doi}{doi: \begingroup \urlstyle{rm}\Url}\fi

\bibitem[Agrawal et~al.(2019)Agrawal, Desai, Wang, Chen, Jain, Johnson, Batra, Parikh, Lee, and Anderson]{Agrawal2019nocaps}
Harsh Agrawal, Karan Desai, Yufei Wang, Xinlei Chen, Rishabh Jain, Mark Johnson, Dhruv Batra, Devi Parikh, Stefan Lee, and Peter Anderson.
\newblock nocaps: novel object captioning at scale.
\newblock \emph{International Conference on Computer Vision}, pages 8947--8956, 2019.

\bibitem[Alayrac et~al.(2022)Alayrac, Donahue, Luc, Miech, Barr, Hasson, Lenc, Mensch, Millican, Reynolds, Ring, Rutherford, Cabi, Han, Gong, Samangooei, Monteiro, Menick, Borgeaud, Brock, Nematzadeh, Sharifzadeh, Binkowski, Barreira, Vinyals, Zisserman, and Simonyan]{alayrac2022flamingo}
Jean-Baptiste Alayrac, Jeff Donahue, Pauline Luc, Antoine Miech, Iain Barr, Yana Hasson, Karel Lenc, Arthur Mensch, Katherine Millican, Malcolm Reynolds, Roman Ring, Eliza Rutherford, Serkan Cabi, Tengda Han, Zhitao Gong, Sina Samangooei, Marianne Monteiro, Jacob Menick, Sebastian Borgeaud, Andrew Brock, Aida Nematzadeh, Sahand Sharifzadeh, Mikolaj Binkowski, Ricardo Barreira, Oriol Vinyals, Andrew Zisserman, and Karen Simonyan.
\newblock Flamingo: a visual language model for few-shot learning.
\newblock In \emph{Advances in Neural Information Processing Systems}, 2022.

\bibitem[Anderson et~al.(2016)Anderson, Fernando, Johnson, and Gould]{anderson2016spice}
Peter Anderson, Basura Fernando, Mark Johnson, and Stephen Gould.
\newblock Spice: Semantic propositional image caption evaluation.
\newblock In \emph{Computer Vision--ECCV 2016: 14th European Conference, Amsterdam, The Netherlands, October 11-14, 2016, Proceedings, Part V 14}, pages 382--398. Springer, 2016.

\bibitem[Anderson et~al.(2017)Anderson, He, Buehler, Teney, Johnson, Gould, and Zhang]{Anderson2017BottomUpAT}
Peter Anderson, Xiaodong He, Chris Buehler, Damien Teney, Mark Johnson, Stephen Gould, and Lei Zhang.
\newblock Bottom-up and top-down attention for image captioning and visual question answering.
\newblock \emph{2018 IEEE/CVF Conference on Computer Vision and Pattern Recognition}, pages 6077--6086, 2017.

\bibitem[Asai et~al.(2024)Asai, Wu, Wang, Sil, and Hajishirzi]{asai2024selfrag}
Akari Asai, Zeqiu Wu, Yizhong Wang, Avirup Sil, and Hannaneh Hajishirzi.
\newblock Self-{RAG}: Learning to retrieve, generate, and critique through self-reflection.
\newblock In \emph{The Twelfth International Conference on Learning Representations}, 2024.

\bibitem[Banerjee and Lavie(2005)]{banerjee2005meteor}
Satanjeev Banerjee and Alon Lavie.
\newblock Meteor: An automatic metric for mt evaluation with improved correlation with human judgments.
\newblock In \emph{Proceedings of the acl workshop on intrinsic and extrinsic evaluation measures for machine translation and/or summarization}, pages 65--72, 2005.

\bibitem[Baranchuk et~al.(2022)Baranchuk, Voynov, Rubachev, Khrulkov, and Babenko]{baranchuk2022ddpmseg}
Dmitry Baranchuk, Andrey Voynov, Ivan Rubachev, Valentin Khrulkov, and Artem Babenko.
\newblock Label-efficient semantic segmentation with diffusion models.
\newblock In \emph{International Conference on Learning Representations}, 2022.

\bibitem[Chiang et~al.(2023)Chiang, Li, Lin, Sheng, Wu, Zhang, Zheng, Zhuang, Zhuang, Gonzalez, et~al.]{chiang2023vicuna}
Wei-Lin Chiang, Zhuohan Li, Zi Lin, Ying Sheng, Zhanghao Wu, Hao Zhang, Lianmin Zheng, Siyuan Zhuang, Yonghao Zhuang, Joseph~E Gonzalez, et~al.
\newblock Vicuna: An open-source chatbot impressing gpt-4 with 90\%* chatgpt quality.
\newblock \emph{See https://vicuna. lmsys. org (accessed 14 April 2023)}, 2\penalty0 (3):\penalty0 6, 2023.

\bibitem[Devlin et~al.(2019)Devlin, Chang, Lee, and Toutanova]{Devlin2019bert}
Jacob Devlin, Ming-Wei Chang, Kenton Lee, and Kristina Toutanova.
\newblock Bert: Pre-training of deep bidirectional transformers for language understanding.
\newblock In \emph{North American Chapter of the Association for Computational Linguistics}, 2019.

\bibitem[Dosovitskiy(2020)]{dosovitskiy2020visiontransformer}
Alexey Dosovitskiy.
\newblock An image is worth 16x16 words: Transformers for image recognition at scale.
\newblock \emph{arXiv preprint arXiv:2010.11929}, 2020.

\bibitem[Fang et~al.(2022)Fang, Wang, Xie, Sun, Wu, Wang, Huang, Wang, and Cao]{Fang2022eva}
Yuxin Fang, Wen Wang, Binhui Xie, Quan-Sen Sun, Ledell~Yu Wu, Xinggang Wang, Tiejun Huang, Xinlong Wang, and Yue Cao.
\newblock Eva: Exploring the limits of masked visual representation learning at scale.
\newblock \emph{2023 IEEE/CVF Conference on Computer Vision and Pattern Recognition (CVPR)}, pages 19358--19369, 2022.

\bibitem[Fei et~al.(2023)Fei, Wang, Zhang, He, Wang, and Zheng]{Fei2023viecap}
Junjie Fei, Teng Wang, Jinrui Zhang, Zhenyu He, Chengjie Wang, and Feng Zheng.
\newblock Transferable decoding with visual entities for zero-shot image captioning.
\newblock \emph{2023 IEEE/CVF International Conference on Computer Vision (ICCV)}, pages 3113--3123, 2023.

\bibitem[Guu et~al.(2020)Guu, Lee, Tung, Pasupat, and Chang]{guu2020rag1}
Kelvin Guu, Kenton Lee, Zora Tung, Panupong Pasupat, and Mingwei Chang.
\newblock Retrieval augmented language model pre-training.
\newblock In \emph{International conference on machine learning}, pages 3929--3938. PMLR, 2020.

\bibitem[Karpathy and Fei-Fei(2014)]{Karpathy2014DeepVA}
Andrej Karpathy and Li Fei-Fei.
\newblock Deep visual-semantic alignments for generating image descriptions.
\newblock \emph{2015 IEEE Conference on Computer Vision and Pattern Recognition (CVPR)}, pages 3128--3137, 2014.

\bibitem[Kim et~al.(2024)Kim, Kim, Moon, Choi, and Kim]{Kim2024cm2}
Minkuk Kim, Hyeon~Bae Kim, Jinyoung Moon, Jinwoo Choi, and Seong~Tae Kim.
\newblock Do you remember? dense video captioning with cross-modal memory retrieval.
\newblock \emph{2024 IEEE/CVF Conference on Computer Vision and Pattern Recognition (CVPR)}, pages 13894--13904, 2024.

\bibitem[Lewis et~al.(2020)Lewis, Perez, Piktus, Petroni, Karpukhin, Goyal, K{\"u}ttler, Lewis, Yih, Rockt{\"a}schel, et~al.]{lewis2020rag2}
Patrick Lewis, Ethan Perez, Aleksandra Piktus, Fabio Petroni, Vladimir Karpukhin, Naman Goyal, Heinrich K{\"u}ttler, Mike Lewis, Wen-tau Yih, Tim Rockt{\"a}schel, et~al.
\newblock Retrieval-augmented generation for knowledge-intensive nlp tasks.
\newblock \emph{Advances in Neural Information Processing Systems}, 33:\penalty0 9459--9474, 2020.

\bibitem[Li et~al.(2022)Li, Li, Xiong, and Hoi]{Li2022BLIPBL}
Junnan Li, Dongxu Li, Caiming Xiong, and Steven C.~H. Hoi.
\newblock Blip: Bootstrapping language-image pre-training for unified vision-language understanding and generation.
\newblock In \emph{International Conference on Machine Learning}, 2022.

\bibitem[Li et~al.(2023{\natexlab{a}})Li, Li, Savarese, and Hoi]{Li2023BLIP2BL}
Junnan Li, Dongxu Li, Silvio Savarese, and Steven C.~H. Hoi.
\newblock Blip-2: Bootstrapping language-image pre-training with frozen image encoders and large language models.
\newblock In \emph{International Conference on Machine Learning}, 2023{\natexlab{a}}.

\bibitem[Li et~al.(2023{\natexlab{b}})Li, Vo, Sugimoto, and Nakayama]{Li2023EvcapRI}
Jiaxuan Li, Duc~Minh Vo, Akihiro Sugimoto, and Hideki Nakayama.
\newblock Evcap: Retrieval-augmented image captioning with external visual-name memory for open-world comprehension.
\newblock \emph{2024 IEEE/CVF Conference on Computer Vision and Pattern Recognition (CVPR)}, pages 13733--13742, 2023{\natexlab{b}}.

\bibitem[Lin et~al.(2014)Lin, Maire, Belongie, Hays, Perona, Ramanan, Doll{\'a}r, and Zitnick]{Lin2014coco}
Tsung-Yi Lin, Michael Maire, Serge~J. Belongie, James Hays, Pietro Perona, Deva Ramanan, Piotr Doll{\'a}r, and C.~Lawrence Zitnick.
\newblock Microsoft coco: Common objects in context.
\newblock In \emph{European Conference on Computer Vision}, 2014.

\bibitem[Liu et~al.(2023)Liu, Li, Fei, Fu, Luo, and Guo]{Liu2023PrefixdiffusionAL}
Guisheng Liu, Yi Li, Zhengcong Fei, Haiyan Fu, Xiangyang Luo, and Yanqing Guo.
\newblock Prefix-diffusion: A lightweight diffusion model for diverse image captioning.
\newblock \emph{ArXiv}, abs/2309.04965, 2023.

\bibitem[Luo et~al.(2024)Luo, Dunlap, Park, Holynski, and Darrell]{luo2024diffusionhyper}
Grace Luo, Lisa Dunlap, Dong~Huk Park, Aleksander Holynski, and Trevor Darrell.
\newblock Diffusion hyperfeatures: Searching through time and space for semantic correspondence.
\newblock \emph{Advances in Neural Information Processing Systems}, 36, 2024.

\bibitem[Luo et~al.(2022)Luo, Li, Pan, Yao, Feng, Chao, and Mei]{Luo2022scdnet}
Jianjie Luo, Yehao Li, Yingwei Pan, Ting Yao, Jianlin Feng, Hongyang Chao, and Tao Mei.
\newblock Semantic-conditional diffusion networks for image captioning*.
\newblock \emph{2023 IEEE/CVF Conference on Computer Vision and Pattern Recognition (CVPR)}, pages 23359--23368, 2022.

\bibitem[Mokady and Hertz(2021)]{Mokady2021ClipCapCP}
Ron Mokady and Amir Hertz.
\newblock Clipcap: Clip prefix for image captioning.
\newblock \emph{ArXiv}, abs/2111.09734, 2021.

\bibitem[Pan et~al.(2023)Pan, Lin, Ge, Zhu, Zhang, Wang, Qiao, and Li]{Pan2023RetrievingtoAnswerZV}
Junting Pan, Ziyi Lin, Yuying Ge, Xiatian Zhu, Renrui Zhang, Yi Wang, Yu~Jiao Qiao, and Hongsheng Li.
\newblock Retrieving-to-answer: Zero-shot video question answering with frozen large language models.
\newblock \emph{2023 IEEE/CVF International Conference on Computer Vision Workshops (ICCVW)}, pages 272--283, 2023.

\bibitem[Papineni et~al.(2002)Papineni, Roukos, Ward, and Zhu]{Papineni2002BleuAM}
Kishore Papineni, Salim Roukos, Todd Ward, and Wei-Jing Zhu.
\newblock Bleu: a method for automatic evaluation of machine translation.
\newblock In \emph{Annual Meeting of the Association for Computational Linguistics}, 2002.

\bibitem[Plummer et~al.(2015)Plummer, Wang, Cervantes, Caicedo, Hockenmaier, and Lazebnik]{Plummer2015Flickr30kEC}
Bryan~A. Plummer, Liwei Wang, Christopher~M. Cervantes, Juan~C. Caicedo, J. Hockenmaier, and Svetlana Lazebnik.
\newblock Flickr30k entities: Collecting region-to-phrase correspondences for richer image-to-sentence models.
\newblock \emph{International Journal of Computer Vision}, 123:\penalty0 74 -- 93, 2015.

\bibitem[Prabhudesai et~al.(2023)Prabhudesai, Ke, Li, Pathak, and Fragkiadaki]{Prabhudesai2023DiffusionTTATA}
Mihir Prabhudesai, Tsung-Wei Ke, Alexander~C. Li, Deepak Pathak, and Katerina Fragkiadaki.
\newblock Diffusion-tta: Test-time adaptation of discriminative models via generative feedback.
\newblock \emph{ArXiv}, abs/2311.16102, 2023.

\bibitem[Ramos et~al.(2022)Ramos, Martins, Elliott, and Kementchedjhieva]{Ramos2022SmallcapLI}
Rita~Parada Ramos, Bruno Martins, Desmond Elliott, and Yova Kementchedjhieva.
\newblock Smallcap: Lightweight image captioning prompted with retrieval augmentation.
\newblock \emph{2023 IEEE/CVF Conference on Computer Vision and Pattern Recognition (CVPR)}, pages 2840--2849, 2022.

\bibitem[Ramos et~al.(2023)Ramos, Elliott, and Martins]{Ramos2023extra}
Rita~Parada Ramos, Desmond Elliott, and Bruno Martins.
\newblock Retrieval-augmented image captioning.
\newblock In \emph{Conference of the European Chapter of the Association for Computational Linguistics}, 2023.

\bibitem[Rombach et~al.(2021)Rombach, Blattmann, Lorenz, Esser, and Ommer]{Rombach2021latentdiffusion}
Robin Rombach, A. Blattmann, Dominik Lorenz, Patrick Esser, and Bj{\"o}rn Ommer.
\newblock High-resolution image synthesis with latent diffusion models.
\newblock \emph{2022 IEEE/CVF Conference on Computer Vision and Pattern Recognition (CVPR)}, pages 10674--10685, 2021.

\bibitem[Sutskever et~al.(2011)Sutskever, Martens, and Hinton]{Sutskever2011rnncap}
Ilya Sutskever, James Martens, and Geoffrey~E. Hinton.
\newblock Generating text with recurrent neural networks.
\newblock In \emph{International Conference on Machine Learning}, 2011.

\bibitem[Vaswani(2017)]{vaswani2017attention}
A Vaswani.
\newblock Attention is all you need.
\newblock \emph{Advances in Neural Information Processing Systems}, 2017.

\bibitem[Vedantam et~al.(2014)Vedantam, Zitnick, and Parikh]{Vedantam2014CIDErCI}
Ramakrishna Vedantam, C.~Lawrence Zitnick, and Devi Parikh.
\newblock Cider: Consensus-based image description evaluation.
\newblock \emph{2015 IEEE Conference on Computer Vision and Pattern Recognition (CVPR)}, pages 4566--4575, 2014.

\bibitem[Wang et~al.(2024)Wang, Sun, Zhang, Tang, Liu, and Wang]{wang2024diffusionhelp}
Wenxuan Wang, Quan Sun, Fan Zhang, Yepeng Tang, Jing Liu, and Xinlong Wang.
\newblock Diffusion feedback helps clip see better.
\newblock \emph{arXiv preprint arXiv:2407.20171}, 2024.

\bibitem[Wang et~al.(2022)Wang, Li, Xu, Zhou, Lei, Lin, Wang, Yang, Zhu, Hoiem, et~al.]{wang2022vidil}
Zhenhailong Wang, Manling Li, Ruochen Xu, Luowei Zhou, Jie Lei, Xudong Lin, Shuohang Wang, Ziyi Yang, Chenguang Zhu, Derek Hoiem, et~al.
\newblock Language models with image descriptors are strong few-shot video-language learners.
\newblock \emph{arXiv preprint arXiv:2205.10747}, 2022.

\bibitem[Yang et~al.(2023)Yang, Ping, Liu, Korthikanti, Nie, Huang, Fan, Yu, Lan, Li, Liu, Zhu, Shoeybi, Catanzaro, Xiao, and Anandkumar]{Yang2023Revilm}
Zhuolin Yang, Wei Ping, Zihan Liu, Vijay~Anand Korthikanti, Weili Nie, De-An Huang, Linxi~(Jim) Fan, Zhiding Yu, Shiyi Lan, Bo Li, Mingyan Liu, Yuke Zhu, Mohammad Shoeybi, Bryan Catanzaro, Chaowei Xiao, and Anima Anandkumar.
\newblock Re-vilm: Retrieval-augmented visual language model for zero and few-shot image captioning.
\newblock In \emph{Conference on Empirical Methods in Natural Language Processing}, 2023.

\bibitem[Yasunaga et~al.(2023)Yasunaga, Aghajanyan, Shi, James, Leskovec, Liang, Lewis, Zettlemoyer, and tau Yih]{Yasunaga2023Racm3}
Michihiro Yasunaga, Armen Aghajanyan, Weijia Shi, Rich James, Jure Leskovec, Percy Liang, Mike Lewis, Luke Zettlemoyer, and Wen tau Yih.
\newblock Retrieval-augmented multimodal language modeling.
\newblock \emph{ArXiv}, abs/2211.12561, 2023.

\bibitem[Zeng et~al.(2024)Zeng, Xie, Zhang, Chen, Wang, and Chen]{Zeng2024MeaCapMZ}
Zequn Zeng, Yan Xie, Hao Zhang, Chiyu Chen, Zhengjue Wang, and Boli Chen.
\newblock Meacap: Memory-augmented zero-shot image captioning.
\newblock \emph{2024 IEEE/CVF Conference on Computer Vision and Pattern Recognition (CVPR)}, pages 14100--14110, 2024.

\bibitem[Zhu et~al.(2023)Zhu, Chen, Shen, Li, and Elhoseiny]{Zhu2023MiniGPT4EV}
Deyao Zhu, Jun Chen, Xiaoqian Shen, Xiang Li, and Mohamed Elhoseiny.
\newblock Minigpt-4: Enhancing vision-language understanding with advanced large language models.
\newblock \emph{ArXiv}, abs/2304.10592, 2023.

\end{thebibliography}
}
\clearpage
\clearpage
\maketitlesupplementary

\section{Implementation Details of Diffusion Guidance}
\label{sec:Imple}
We adopt the latent diffusion model Stable Diffusion 1-4~\cite{Rombach2021latentdiffusion} to implement diffusion guidance, enabling computationally efficient denoising by operating on latent features rather than the original image resolution. This approach significantly reduces memory and processing requirements while preserving critical visual information. For each training sample, the denoising process is executed once, with the noise level \( t \) randomly selected to simulate a diverse range of diffusion scenarios. The latent features are perturbed according to the chosen noise level, and the resulting noised features are input to a pretrained UNet model, conditioned on the image features, to predict the added noise. By minimizing the difference between the predicted and true noise through the diffusion loss, the image encoder learns to generate robust and comprehensive features that better represent fine-grained visual details. Our single-step denoising strategy optimizes the image encoder with minimal computational cost while maintaining feature expressiveness. The improved encoder enhances the retrieval process by producing compact and informative representations, which in turn support the generation of accurate and contextually rich captions.


\section{Construction of Retrieval Database}
The retrieval database is constructed using the training set of the COCO~\cite{Lin2014coco} dataset. For each image, we extract image features using EVA-CLIP~\cite{Fang2022eva}, which maps images and text to a shared embedding space. To enrich contextual information, we parse all five ground-truth captions associated with each image to extract key attributes such as objects, actions, and environments. This parsing process, which can be performed offline, utilizes the spaCy NLP tool to classify words into categories: ``NOUN" or ``PROPN" for objects, ``VERB" for actions, and ``GPE" or ``LOC" for environments. Non-informative words, such as auxiliary verbs like ``be," are manually filtered out, and words are lemmatized to their root forms (e.g., ``helps" becomes ``help," ``tables" becomes ``table") to ensure consistency. The resulting structured attributes, including objects, actions, and environments, are stored in the retrieval database. This enables better alignment between visual and textual information, ultimately enhancing retrieval effectiveness and improving caption generation.

\section{Additional Ablation Analysis}

\begin{figure}
    \centering
    \includegraphics[width=\linewidth]{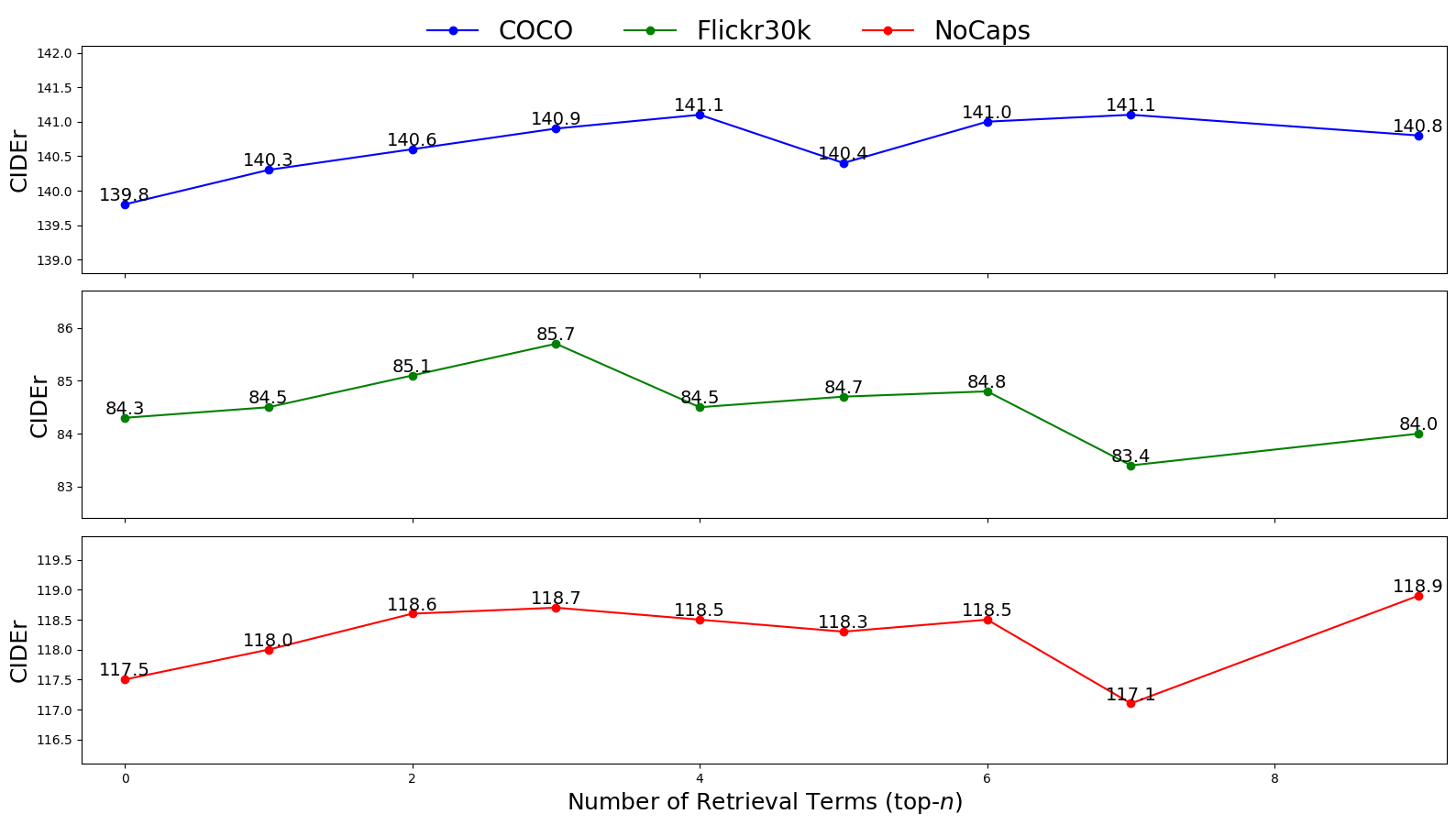}
    \caption{Comparison of model performance with varying top-$n$ values on the CIDEr metric across COCO, Flickr30k, and the out-of-domain subset of NoCaps.}
    \label{fig:topn_term}
\end{figure}

\noindent\textbf{Effect of the Number of Retrieved Terms (top-$n$)}
To investigate the impact of the number of retrieved terms (top-$n$) on model performance, we visualize the CIDEr metric across three datasets: COCO, Flickr30k, and the out-of-domain subset of NoCaps, as shown in \Cref{fig:topn_term}. While the SPICE scores remain similar across different top-$n$ values, the CIDEr scores exhibit more noticeable variation. When top-$n$ is set to 1, model performance decreases, indicating that limited context retrieved from the database hampers the model’s ability to generate accurate captions. As the top-$n$ value increases, performance improves, especially at the beginning, as more context is provided. However, beyond a certain threshold, the performance improvement slows down and may even drop, suggesting that including less relevant terms could introduce noise. A top-$n$ value of 3 provides a good balance, yielding strong performance across all three datasets. Based on these findings, we select a top-$n$ value of 3 to report our model’s performance.

\noindent\textbf{Effect of Retrieval Database Size.}
We evaluate the effect of retrieval database size on model performance by conducting experiments with sizes of 25\%, 50\%, 75\%, and 100\%. As shown in \Cref{tab:effetc_retrieval_size},  increasing the database size generally leads to performance improvements, especially for out-of-domain data. A larger retrieval database offers more diverse and richer contextual information, thereby enhancing the model's ability to generalize across various scenarios. However, performance gains are not always consistent. This variability is also observed in the SmallCap and EVCap, where we hypothesize that uneven data distribution during subset selection may contribute to the inconsistency. Nevertheless, the overall trend indicates that a larger, more comprehensive retrieval database generally improves the model’s performance.

\begin{table}[tb]
\resizebox{\columnwidth}{!}{%
\begin{tabular}{c|cc|cccccc}
\toprule
\multirow{2}{*}{\textbf{Percent}} & \multicolumn{2}{c|}{\textbf{Flickr30k}}  & \multicolumn{2}{c}{\textbf{NoCaps-Near}} & \multicolumn{2}{c}{\textbf{NoCaps-Out}} & \multicolumn{2}{c}{\textbf{NoCaps-All}} \\ \cmidrule(l){2-9} 
                                   & \multicolumn{1}{|c}{C} & S  & \multicolumn{1}{|c}{C} & S & \multicolumn{1}{|c}{C} & S & \multicolumn{1}{|c}{C} & S \\ \midrule
25\%         &  83.3   & 17.9 & 119.0 & 15.7   &   116.4 & 14.9  & 118.9 & 15.4  \\
50\%         &  84.3   & \textbf{18.3}  & 120.1 & \textbf{15.9} &   117.5 & \textbf{15.1}  & 119.9 & \textbf{15.6}  \\
75\%         &  84.5   & 18.2  & 119.8 & 15.8 &   118.3 & 15  & 119.4 & 15.5  \\
100\%       &  \textbf{85.7}  & 18.2 & \textbf{120.6} & 15.8 &  \textbf{118.7} & \textbf{15.1}  & \textbf{120.7} & 15.5  \\ \bottomrule
\end{tabular}%
}
\caption{Comparison of model performance with varying retrieval database sizes on Flickr30k and NoCaps.}
\label{tab:effetc_retrieval_size}
\end{table}

\paragraph{Effect of Language Decoder}
We conduct an ablation study to evaluate the effect of the language decoder size on model performance. As shown in \Cref{tab:llm_decoder}, models using larger language decoders consistently achieve better results across all metrics, particularly on the out-of-domain subset of NoCaps. This improvement can be attributed to the enhanced reasoning and generalization capabilities of larger language models, which are better equipped to handle diverse and complex scenarios. The increased model capacity allows for more effective integration of the retrieved contextual information, enabling the generation of richer and more accurate captions, especially for challenging out-of-domain data.
\begin{table}[tb]
\resizebox{\columnwidth}{!}{%
\begin{tabular}{c|cc|cccccc}
\toprule
\multirow{2}{*}{\textbf{LLM}} & \multicolumn{2}{c|}{\textbf{Flickr30k}}  & \multicolumn{2}{c}{\textbf{NoCaps-Near}} & \multicolumn{2}{c}{\textbf{NoCaps-Out}} & \multicolumn{2}{c}{\textbf{NoCaps-All}} \\ \cmidrule(l){2-9} 
                                   & \multicolumn{1}{|c}{C} & S  & \multicolumn{1}{|c}{C} & S & \multicolumn{1}{|c}{C} & S & \multicolumn{1}{|c}{C} & S \\ \midrule
  Vicuna-7B       &  84.2   & 18.1  & 118.6 & 15.5 &   112.7 & 14.6  & 117.6 & 15.3  \\
Vicuna-13B       &  \textbf{85.7}  & \textbf{18.2} & \textbf{120.6} & \textbf{15.8} &  \textbf{118.7} & \textbf{15.1}  & \textbf{120.7} & \textbf{15.5}  \\ \bottomrule
\end{tabular}%
}
\caption{Performance comparison of models using different language decoders on Flickr30k and NoCaps datasets.}
\label{tab:llm_decoder}
\end{table}

\section{Visualization Results}

\begin{figure}
        \centering
        \begin{subfigure}[b]{0.95\columnwidth}
            \centering
            \includegraphics[width=\textwidth]{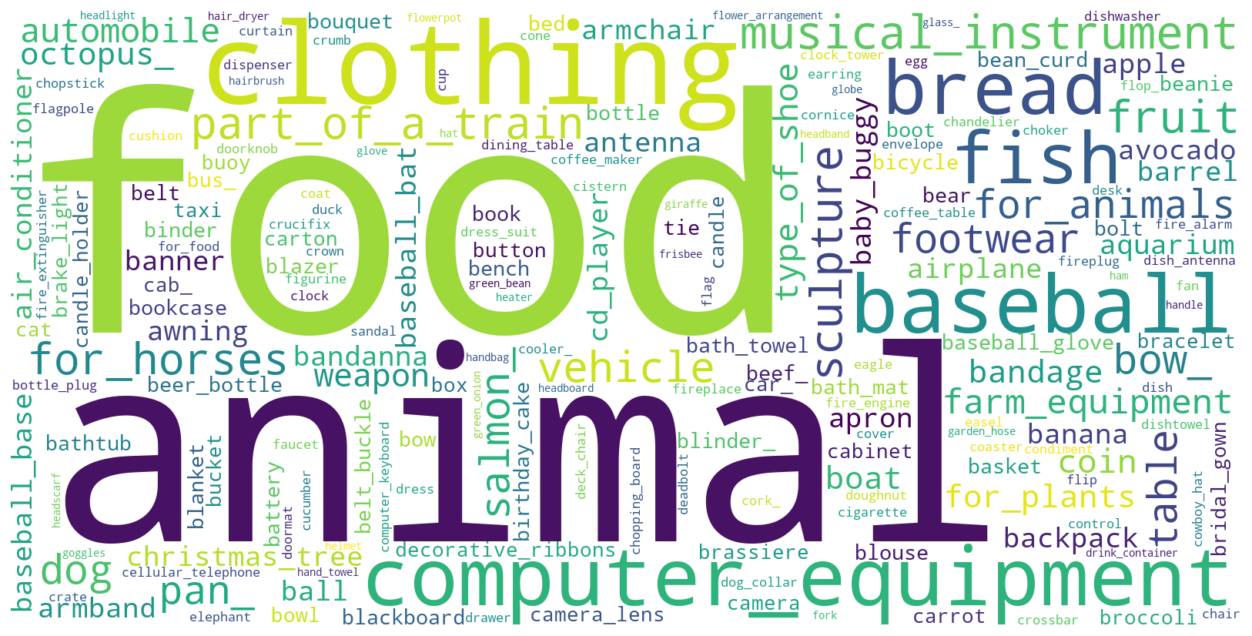}
            \caption{EVCap's Databse}
            \label{fig:wordcloud_evcap}
        \end{subfigure}
        \hfill
        \begin{subfigure}[b]{0.95\columnwidth}  
            \centering 
            \includegraphics[width=\textwidth]{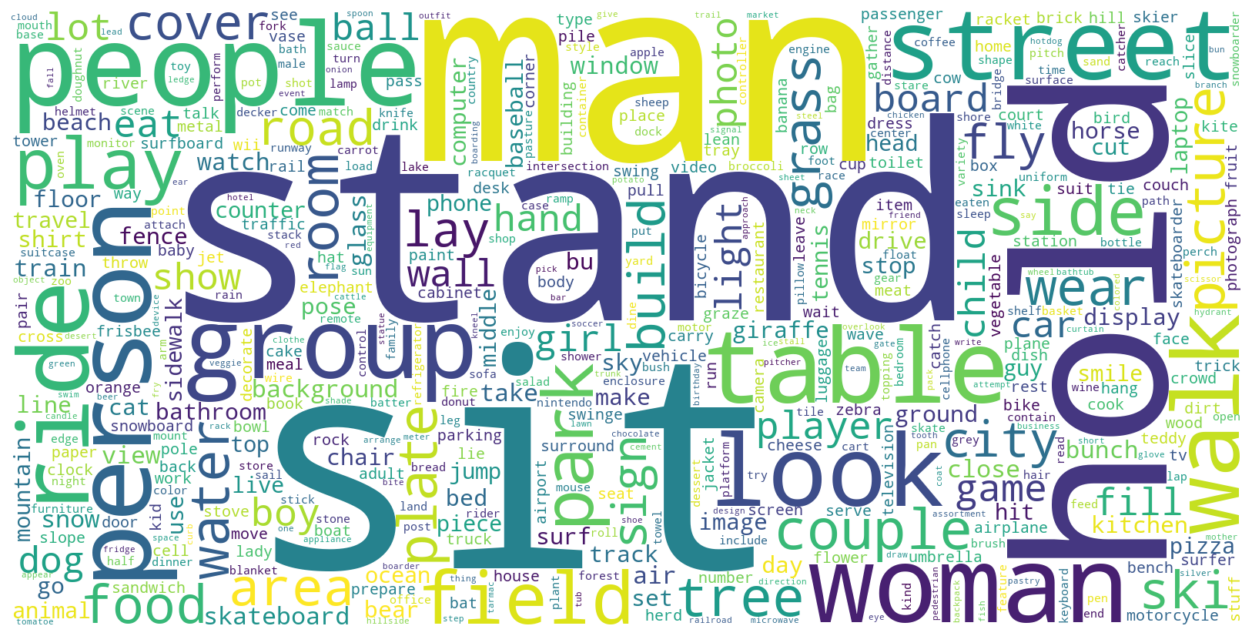}
            \caption{Our Database}   
            \label{fig:wordcloud_ca_object}
        \end{subfigure}
        \caption{Comparison of word clouds illustrating differences between the databases. (a) EVCap's database focusing on objects. (b) Our database's object vocabulary. (c) Actions from our database. (d) Environmental contexts in our database. Our broader retrieval scope enhances contextual richness in generated captions compared to EVCap.}
    \label{fig:wordcloud_comparison}
    \end{figure}

\paragraph{Comparison of Retrieval Databases}
We analyze the word diversity of the retrieval databases used in our model and EVCap by comparing their respective word clouds, as illustrated in \Cref{fig:wordcloud_comparison}. The word cloud generated from our retrieval database demonstrates significantly greater diversity, encompassing a wider range of attributes such as actions, environments, and contextual details. In contrast, the word cloud for EVCap's retrieval database shows a narrower focus, predominantly centered on object-related terms. This comparison highlights the advantage of our retrieval memory, which captures a richer and more balanced representation of visual content, thereby contributing to the generation of more contextually comprehensive captions.

\begin{figure*}[tb]
    \centering
    \includegraphics[width=\linewidth]{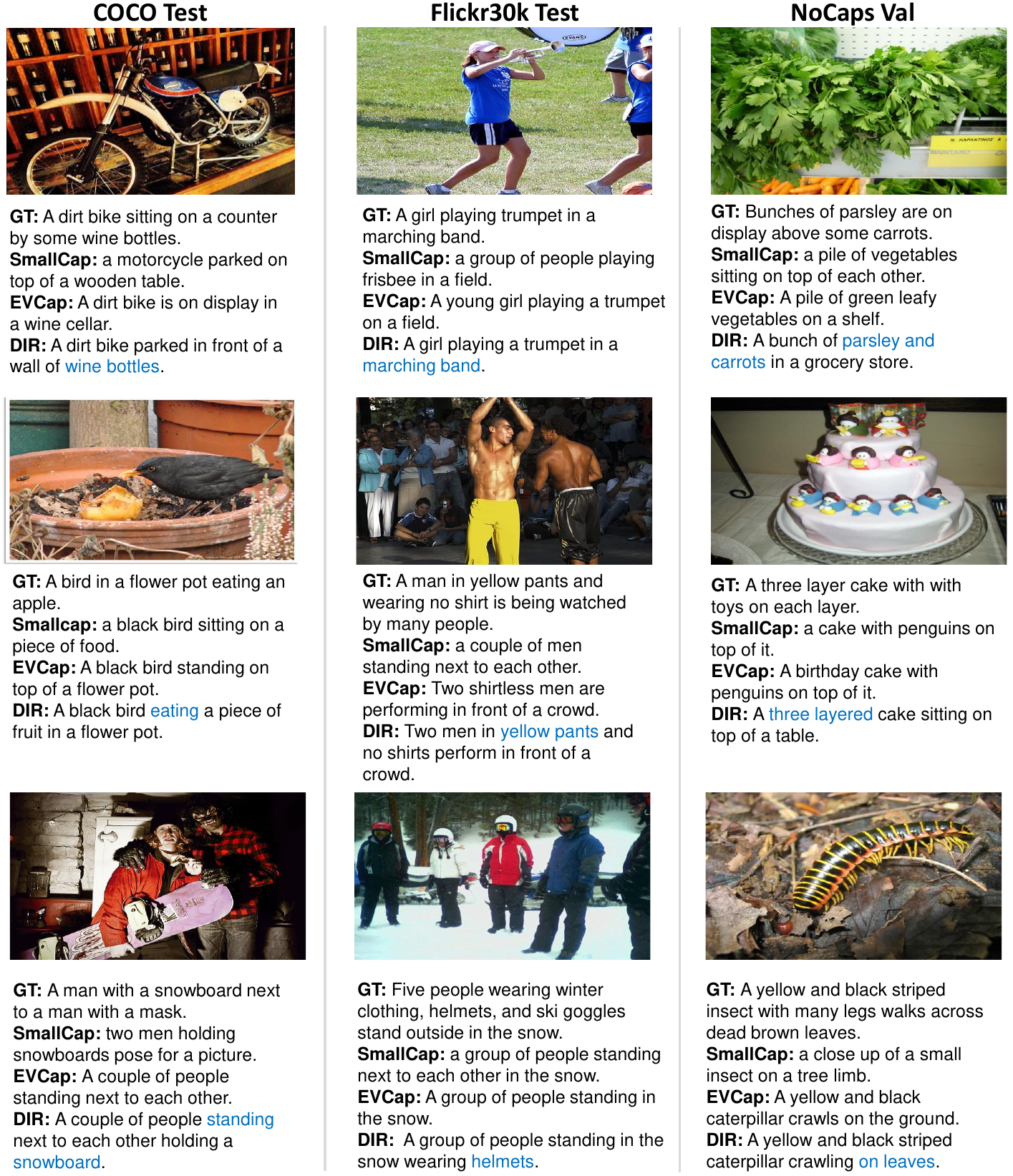}
    \caption{Qualitative comparison of captions generated by SmallCap, EVCap, and our proposed model on COCO, Flickr30k, and NoCaps datasets. The blue text highlights aspects where our model's predictions outperform those of other methods.}
    \label{fig:qualitative_examples}
\end{figure*}

\paragraph{Qualitative Results}
To further evaluate the effectiveness of our approach, we present qualitative comparisons of captions generated by SmallCap, EVCap, and our model, as shown in \Cref{fig:qualitative_examples}. The results show that our model not only maintains strong performance on in-domain datasets but also significantly outperforms on out-of-domain datasets. Captions generated by our model often capture nuanced visual elements, including fine-grained details such as subtle actions and environmental context, providing a more comprehensive representation of the image.

%
%


\end{document}